\documentclass[10pt,twocolumn,letterpaper]{article}

\usepackage{iccv}
\usepackage{times}
\usepackage{epsfig}
\usepackage{graphicx}
\usepackage{amsmath}
\usepackage{amssymb}

\usepackage{ dsfont }
\usepackage{cuted}
\usepackage{capt-of}
\usepackage{verbatim} 
\usepackage[symbol]{footmisc}

\usepackage[pagebackref=true,breaklinks=true,letterpaper=true,colorlinks,bookmarks=false]{hyperref}

\iccvfinalcopy 

\def\iccvPaperID{4230} 

\ificcvfinal\fi

\begin{document}

\title{3D human tongue reconstruction from single ``in-the-wild'' images}

\author{Stylianos Ploumpis$^{1,2}$ \thanks{Authors contributed equally.} 
        \hspace{0.8cm}
        Stylianos Moschoglou$^{1,2}$ \textsuperscript{\thefootnote}  
        \hspace{0.8cm}
        Vasileios Triantafyllou$^2$\\
        \hspace{0.8cm}
        Stefanos Zafeiriou$^{1,2}$ \and
$^1$Imperial College London, UK
\hspace{0.8cm}
$^2$Huawei Technologies Co. Ltd
\\
{\tt\footnotesize $^1$\{s.ploumpis,s.moschoglou,s.zafeiriou\}@imperial.ac.uk}
\hspace{0.5cm}
{\tt\footnotesize $^2$\{vasilios.triantafyllou\}@huawei.com}
}

\maketitle

\begin{strip}\centering
\includegraphics[width=0.9\textwidth]{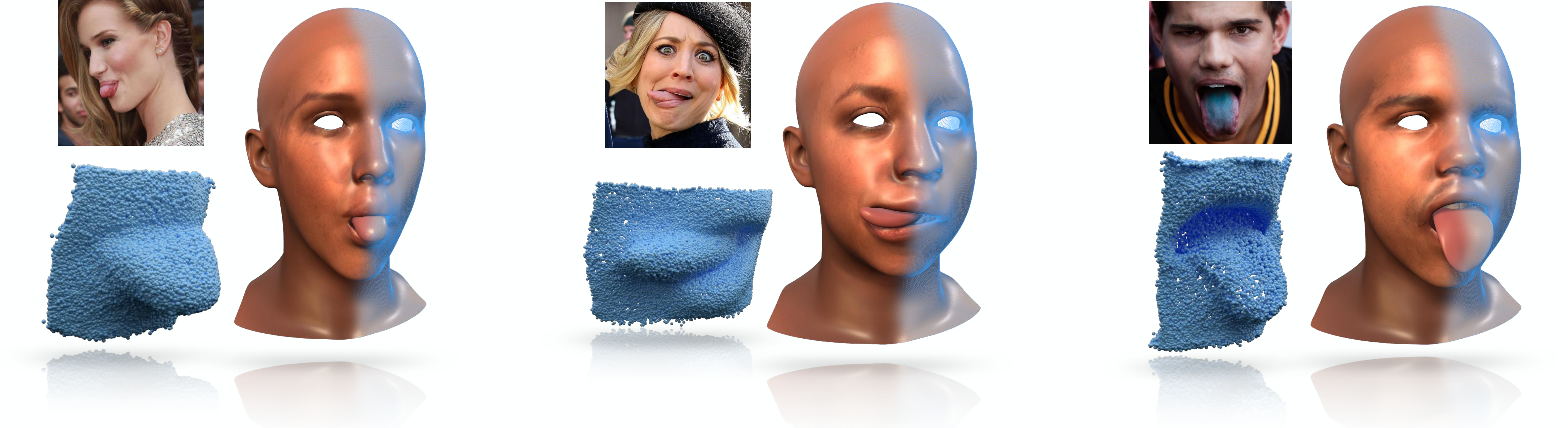}
\captionof{figure}{\small{We propose a framework that accurately derives the 3D tongue shape from single images. A high detailed 3D point cloud of the tongue surface and a full head topology along with the tongue expression can be estimated from the image domain. As we demonstrate, our framework is able to capture the tongue shape even in adverse ``in-the-wild'' conditions.}
\label{fig:feature-graphic}}
\end{strip}
\begin{abstract}
3D face reconstruction from a single image is a task that has garnered increased interest in the Computer Vision community, especially due to its broad use in a number of applications such as realistic 3D avatar creation, pose invariant face recognition and face hallucination. Since the introduction of the 3D Morphable Model in the late 90’s, we witnessed an explosion of research aiming at particularly tackling this task. Nevertheless, despite the increasing level of detail in the 3D face reconstructions from single images mainly attributed to deep learning advances, finer and highly deformable components of the face such as the tongue are still absent from all 3D face models in the literature, although being very important for the realness of the 3D avatar representations. In this work we present the first, to the best of our knowledge, end-to-end trainable pipeline that accurately reconstructs the 3D face together with the tongue. Moreover, we make this pipeline robust in ``in-the-wild'' images by introducing a novel GAN method tailored for 3D tongue surface generation. Finally, we make publicly available to the community the first diverse tongue dataset, consisting of $1,800$ raw scans of $700$ individuals varying in gender, age, and ethnicity backgrounds \footnote{Project url: \url{https://github.com/steliosploumpis/3D_human_tongue_reconstruction}}. As we demonstrate in an extensive series of quantitative as well as qualitative experiments, our model proves to be robust and realistically captures the 3D tongue structure, even in adverse ``in-the-wild'' conditions. 
\end{abstract}

\section{Introduction}
\label{sec:intro}
Recently, 3D face reconstruction from single ``in-the-wild'' images has been a very active topic in Computer Vision with applications ranging from realistic 3D avatar creation to image imputation and face recognition \cite{wu2020unsupervised, gecer2019ganfit, tewari2018self, lattas2020avatarme, ruiz2020morphgan, egger20203d}. Nevertheless, despite the improvement in the quality of the 3D reconstructions, all of these methods do not accommodate any statistical variations in the oral cavity let alone a tongue template mesh. As a result, the oral region is completely disregarded from the final result.

Being able to reconstruct the tongue expression has multiple advantages in various applications. First of all, the generated avatars would be more realistic and would be able to mimic many more facial expressions. Moreover, speech animation tasks would be improved as the inclusion of the oral cavity plays a significant role. Finally, face recognition applications could be enhanced as more extreme poses and expressions would be modeled.

However, as we already pointed out, all of the current state-of-the-art (SOTA) methods \cite{wu2020unsupervised, gecer2019ganfit, tewari2018self} do not contain the tongue component in their implementations. This is because of two reasons: a) there is no publicly available tongue dataset, and b) it is very challenging to carry out 3D reconstruction of the face together with the tongue in ``in-the-wild'' conditions, because of the highly deformable nature of the human tongue.

To tackle the absence of tongue data, we collected a large and diverse dataset of textured 3D tongue point-clouds (more info about the data in Section \ref{sec:tongue_recon}). Having captured the data, we created a pipeline which is comprised of the following parts: a) a tongue point-cloud autoencoder (AE) which is used to derive useful 3D features of our raw collected 3D data, b) a tongue image encoder optimized based on the aforementioned 3D features, c) a shape decoder which translates the encoder outputs to the parameter space of the Universal Head Model (UHM) \cite{ploumpis2020towards}. We should note that the UHM in our case is further rigged/modified so that it can model various tongue shapes/expressions, as explained in Section \ref{sec:tongue_recon}. We begin by training the AE in step a) and then we train steps b-d) in an end-to-end fashion so that the output tongue expression of the UHM model is as close as possible to the corresponding ground-truth 3D tongue point-cloud of the 2D tongue image.

Since there is a lack of ground-truth 3D tongue data corresponding to ``in-the-wild'' 2D tongue images, the pipeline we described so far is only trained using our collected data which were captured under controlled conditions. This results in sub-optimal performance in ``in-the-wild'' conditions. To remedy this, we developed a novel conditional GAN framework that is able to generate accurate 3D tongue point-clouds based on the image encoder outputs (step b) of the pipeline). Having created new image/point-cloud pairs of ``in-the-wild'' tongue data, we re-train the pipeline using also these new data. As we show in Section \ref{sec:experiments}, this addition substantially improves the quality of the tongue reconstructions. 

To summarize, the contributions of our work are the following:

\begin{itemize}
    \item We release a dataset of $1,800$ raw tongue scans of various shapes and positions, corresponding to around $700$ subjects. Being the first such diverse tongue dataset, it can be proven very useful to the community.
    \item We present a complete pipeline trained in an end-to-end fashion that is able to reconstruct the 3D face together with the tongue from a single image.
    \item To make this pipeline robust to ``in-the-wild'' images, we introduce a novel GAN framework which is able to accurately reconstruct 3D tongues from ``in-the-wild'' images with an increasing level of detail.
\end{itemize}

\begin{figure*}[t]
\begin{center}
\includegraphics[width=1\linewidth]{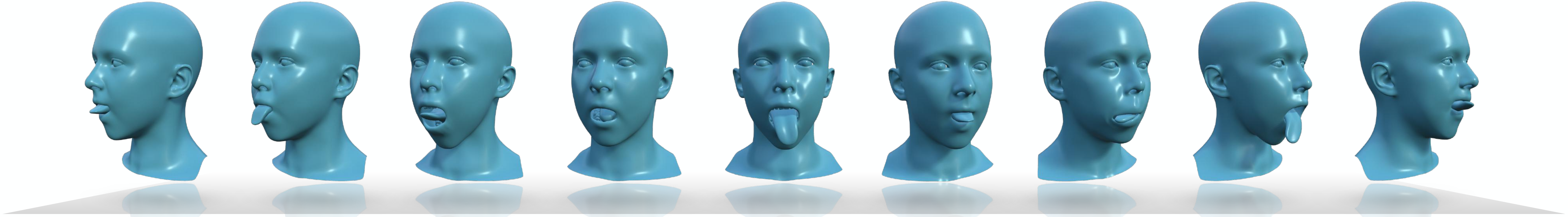}
\caption{Random 3D tongue expressions of our synthetic database based on the mean UHM template. The expressions are rigged and manually sculpted to induce more variance around the tongue surface and the general oral cavity.}
\label{fig:synthetic_data}
\end{center}
\end{figure*}

\section{Related Work}
\label{sec:Related_work}
Single view 3D reconstruction of human facial/head parts is undeniably an extremely valuable task in Computer Vision. However, it has posed many challenges to the research community, due to the fundamental depth ambiguities and the ill-posed nature of the problem. In order to constrain the ambiguity of the problem, many statistical parametric models have been introduced for different parts of the human face/head \cite{blanz1999morphable,booth20163d,li2017learning,ploumpis2019combining}.

Due to the increasing interest of facial analysis over the years the research community has mainly focused on human facial reconstructions. Since the inception of facial 3D Morphable Models (3DMMs) in \cite{blanz1999morphable}, a myriad of scientific papers have been published focusing solely on the reconstruction of facial shape and appearance \cite{booth20173d,booth20183d,gecer2019ganfit,lattas2020avatarme}. Only recently with the emergence of 3D scanning data has the research interest shifted to other significant parts of the human head. A few head models have been introduced during the recent years but without any statistical craniofacial correlations \cite{li2017learning, ranjan2018generating}. The first craniofacial 3DMM of the human head was introduced  in \cite{dai20173d} and later extended and leveraged into a 3D head reconstruction setting from unconstrained single images \cite{ploumpis2019combining}. A few recent works tried to align a skull structure of the the human head with the facial topology \cite{liu2018superimposition, madsen2018probabilistic} in order to obtain a distribution of plausible face shapes given a skull shape.

Finer details of the human face/head started to appear with the introduction of 3D human ear modeling \cite{zhou2017deformable}. Ears are key structures of the human head that have an important contribution to the biometric recognition and general appearance of a person. The two foremost examples of ear models were introduced in \cite{zolfghari2016ear,dai2018ear} but none of them was fused to a face/head in order to create a complete appearance.

Moreover, in an attempt to overcome the ``\emph{uncanny valley}'' problem, a few approaches have tried to model the independent variations/movements of the human eye and the the facial eye region \cite{wood20163d, berard2016lightweight}. These efforts are challenging due to the limited amount of data around the eye region and the extreme level of detail required for this task. Moving towards the oral cavity, teeth modeling was introduced in \cite{wu2016model,velinov2018appearance}, where the 3D structure of the teeth was recovered from 2D images via an elaborate optimization scheme. 

Only very recently, a few approaches \cite{li2020learning, ploumpis2020towards} have tried to combine all of these aforementioned attributes of the human head (eyes, ears, teeth, and inner oral cavity) in order to build a complete model in terms of shape and texture, which accurately represents the human head. Although these models include an oral topology, none of them deals with the dynamics of the tongue, something which is really important for speech animation and the overall realness of the avatar representation. To this end, in this work we aim at extending these approaches and paving the way towards a realistic human appearance by releasing a diverse 3D tongue dataset to the research community. We also present the first framework for accurate 3D human tongue reconstruction from single images.


\begin{figure*}[t]
\begin{center}
\includegraphics[width=0.90\linewidth]{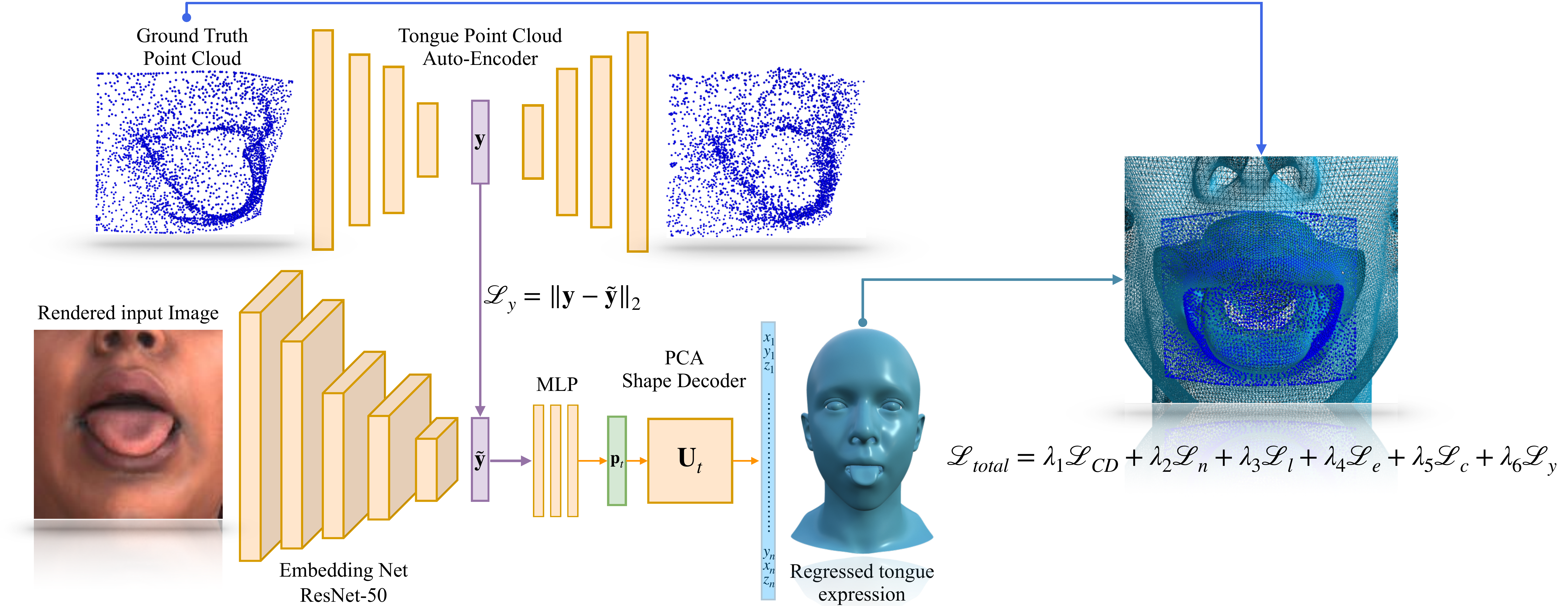}
\caption{An illustration of our tongue reconstruction framework. First we train the point-cloud AE on its own to get meaningful 3D features ($\mathbf{y}$) and then we trained the image encoder /shape decoder using a number of different losses as explained in Section \ref{sec:tongue_recon}}
\label{fig:method}
\end{center}
\end{figure*}

\section{3D human tongue reconstruction}
\label{sec:tongue_recon}
In this Section we present the complete tongue reconstruction pipeline. We begin by describing our collected 2D/3D tongue dataset and our manually rigged tongue dataset which is based on the the UHM \cite{ploumpis2020towards} template. We further provide details about the point-cloud AE, the image encoder, the shape decoder and the overall loss functions we used to optimize the pipeline for the tongue reconstruction. Moreover, we present the novel conditional GAN method which is able to accurately reconstruct 3D tongue point-clouds of ``in-the-wild'' tongue images. Finally, we explain how we used the generated point-clouds of the GAN to re-train the pipeline to achieve better results in ``in-the-wild'' conditions.

\subsection{Tongue datasets}
\label{sec:datasets}
\begin{figure}[t]
\begin{center}
\includegraphics[width=0.95\linewidth]{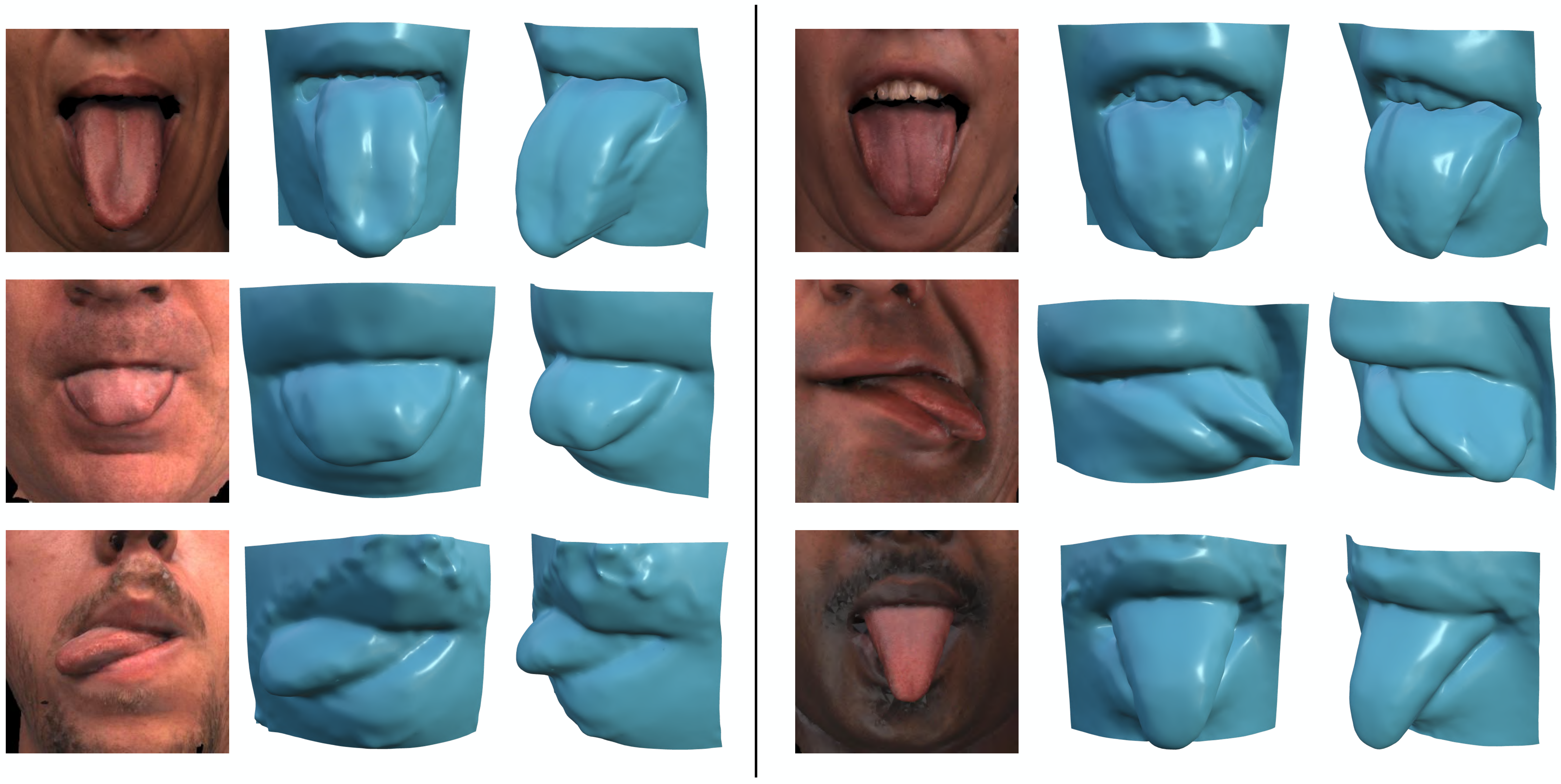}
\caption{Various raw 3D tongue scans of our database depicting different tongue expressions along with the corresponding 2D renders.}
\label{fig:raw_dataset}
\end{center}
\end{figure}

\textbf{TongueDB: the first 3D tongue dataset.} 
As mentioned in Section \ref{sec:intro}, we collected a large dataset comprising of textured 3D tongue scans. Our point cloud database, dubbed TongueDB, contains approximately 1,800 3D tongue scans which were captured during a special exhibition in the Science Museum, London. The subjects were instructed to perform a range of tongue expressions (\ie, tongue out left and right, tongue out center, tongue out center round, tongue out center extreme open mouth, tongue inside left and right, etc.). Some example images can be seen in Fig. \ref{fig:raw_dataset}. The capturing apparatus utilized for this task was a 3dMD 4 camera structured light stereo system, which produces high quality dense meshes. We recorded a total of 700 distinct subjects with available metadata about them, including their gender ($42\%$ male, $58\%$ female), age, and ethnicity ($82\%$ White, $9\%$ Asian, $3\%$ Black and $6\%$ other).

\textbf{Rigged tongue database.}
In order to carry out 3D tongue and face reconstruction, we would need to use a face/head model. Nevertheless, one major drawback of all of the currently used face/head models \cite{li2017learning,ranjan2018generating,dai20173d} is that they are missing the tongue component. This is because it is a challenging task to non-rigidly capture in a fixed template the 3D topology of the oral cavity. These challenges include: a) the highly deformable nature of the tongue, b) the non-convexity of the mouth region, c) the specular texture of the teeth. In order to alleviate this issue, we constructed a synthetic 3D head and tongue dataset rigged by 3D artists. For our neutral mesh template $\mathbf{\bar{T}}$ we utilize the \textsl{mean template} of the UHM \cite{ploumpis2020towards} as it provides all the necessary components of the human oral cavity in accordance with the entire head statistical structure. The resulting rigged tongue expressions rise at $75$ distinct meshes. In order to further augment our synthetic dataset, we performed trilinear interpolation between the closest expression meshes and generated a total of $n_s=720$ tongue expressions. Some example synthetic expressions are shown in Fig. \ref{fig:synthetic_data}. A standard PCA was applied on the interpolated meshes resulting in an orthogonal basis matrix $\mathbf{U}_t \in \mathds{R}^{3N\times n_t}$ (where $N$ are the mesh vertices and $n_t=110$ the kept components). The PCA is performed on the entire set of head vertices and not solely on the oral cavity. In this way, it is more efficient afterwards to transfer the tongue expression from the mean head to a head with a different facial identity.


\subsection{Method}
\label{sec:method}
\textbf{Tongue point-cloud AE.}
In order to accurately reconstruct a tongue in its 3D form based on a 2D image, our image encoder needs to be guided by meaningful target labels which can capture all the desired 3D point-cloud information. These labels, denoted as $\mathbf{y}\in\mathbb{R}^{256}$, are learned by autoencoding the raw point-clouds of our dataset (\ie, the raw 3D tongue scans). For this task, we utilize a self organizing-map framework for hierarchical feature extraction \cite{li2018so}.

\textbf{Tongue image encoder.}
\label{sec:emb_net}
The task of the tongue image encoder is to produce features which are close to the target 3D features $\mathbf{y}$ of the AE. To make the encoder robust to various camera angles or illuminations, we employ a rendering framework where we utilize the textured raw scans (TongueDB in Section \ref{sec:datasets}). We render our 1,8K textured meshes with a pre-computed radiance transfer technique using spherical harmonics which efficiently represent global light scattering. Additionally, we use more than 15 different indoor scenes coupled with random light positions and mesh orientations around all 3D axes, resulting in approximately 100K images. As an encoder we used a ResNet-50 \cite{he2016deep} model pre-trained on ImageNet \cite{deng2009imagenet} and fine-tuned it on our dataset. In particular, we modified the last layer of the network to output a vector $\tilde{\mathbf{y}} \in \mathbb{R}^{256}$ similar to the dimension of the ground truth vector $\mathbf{y}$.

\textbf{Shape decoder.}
In order to decode the encoder $\tilde{\mathbf{y}}$ labels into meaningful tongue shapes, we use the synthetic PCA model $\mathbf{U}_t$ of the rigged tongue expression dataset. To this end, after producing the $\tilde{\mathbf{y}}$ labels, we utilize a standard multi-layer perceptron (MLP) which works as a regression scheme to the latent parameters $\mathbf{p}_t \in \mathds{R}^{110}$ of the synthetic PCA tongue model. The statistical nature of the PCA model helps us constrain the final result during training and ensures meaningful deformations which lie inside the spectrum of our rigged/modified UHM model.

\textbf{Pipeline training}.
During training, we first train the point-cloud AE on its own and then train the pipeline of the image encoder and shape decoder in an end-to-end fashion. To optimize the pipeline, we apply a total of $6$ losses with each one contributing to the quality of the final result. The first $2$ losses are calculated between the predicted tongue expression of the rigged/modified UHM model and the ground-truth tongue point-cloud of the corresponding input image. Similarly to \cite{wang2018pixel2mesh}, we adopt a Chamfer loss \cite{fan2017point} $\mathcal{L}_{CD}$ to optimize the position of the resulting template points as well as a normal loss $\mathcal{L}_{n}$ to correct the orientation of the mesh. In order to compute an accurate Chamfer loss, we only utilize a small area around the oral cavity which is defined based on the ground-truth point-cloud. Additionally, we calculate a Laplacian regularization $\mathcal{L}_{l}$ loss between our predicted mesh the mean shape of the PCA model in order to prevent the vertices from moving too freely outside the mean positions and constrain the resulting shape to be smooth. An edge length loss $\mathcal{L}_{e}$ is also introduced which penalizes any flying vertices (outliers). Finally, we employ a collision loss $\mathcal{L}_{c}$  which prevents the points of the tongue to penetrate the surface of the oral cavity and is formulated as the sum of each collision error around the $12$ mouth landmarks of the UHM template (as illustrated in the supplementary material):

\begin{equation}
\mathcal{L}_{col}=\frac{1}{N} \sum_{k=0}^{11}\sum_{i=0}^{N-1}\max\left(0, d_{k}^i\right)
\end{equation}
\[d_{k}^i = r^2 - \left(q_{1}^i - x_k\right)^2 - \left(q_{2}^i - y_k\right)^2 - \left(q_{3}^i - z_k\right)^2\]

\begin{figure*}[h!]
\begin{center}
\includegraphics[width=1\linewidth]{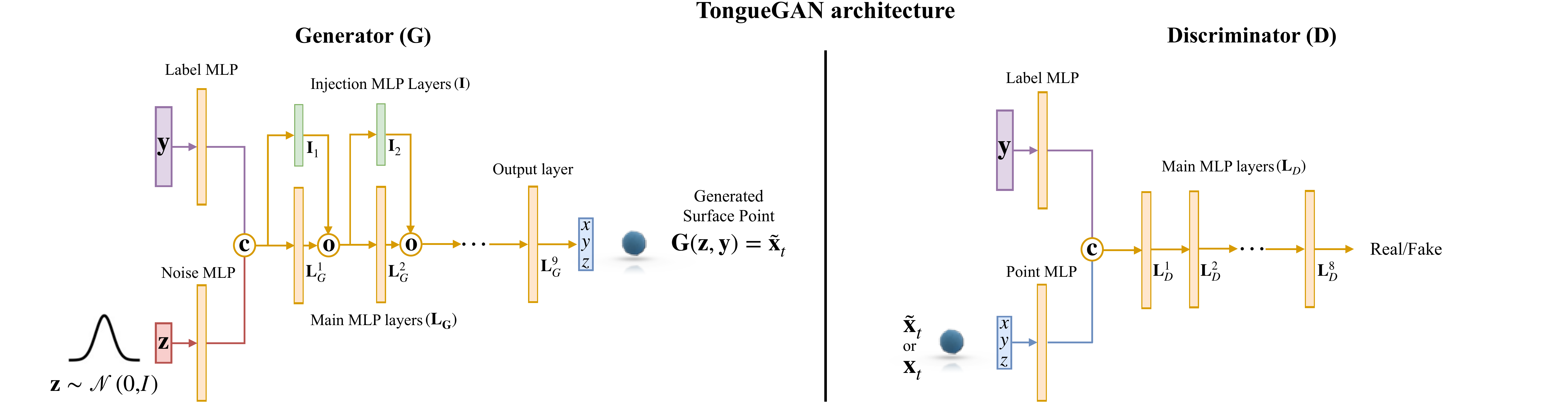}
\caption{Symbol $\mathbf{c}$ stands for row-wise concatenation along the channel dimension. Symbol $\mathbf{o}$ stands for element-wise (\ie, Hadamard) product. The Generator inputs are a Gaussian noise sample $\mathbf{z}$ and a label $\mathbf{y}$ corresponding to a particular tongue, from which we want to sample a 3D point. The Discriminator input pairs are a label $\mathbf{y}$ which corresponds to a specific tongue and $\mathbf{x}_t$ a \textsl{real} 3D point belonging to the aforementioned tongue point-cloud (but sampled as explained in Section \ref{sec:novel_loss}) and $G\left(\mathbf{z}, \mathbf{y}\right) = \tilde{\mathbf{x}_t}$ a generated point belonging to this tongue. The Discriminator is asked to distinguish the real from the fake point.}
\label{fig:net_diagram}
\end{center}
\end{figure*}

The $\mathcal{L}_{col}$ is calculated as the sum of distances of each collided point $\mathbf{q}^i=\{q_{1}^i,q_{2}^i,q_{3}^i\}$ to the sphere $k$ with center the landmark coordinates $x_k, y_k, z_k$ and radius $r=1,5cm$.

Lastly, we impose a final L2 loss $\mathcal{L}_{y}$ in the intermediate step of our pipeline where we constrain the predicted $\tilde{\mathbf{y}}$ encoded features to be as close as possible to the ground-truth features ${\mathbf{y}}$ of the corresponding autoencoded point-cloud. This loss is of paramount importance because: a) the $\tilde{\mathbf{y}}$ features in this way contain rich 3D information invariant to texture/illumination variations and b) our ``in-the-wild'' extension which we introduce later is based on a generative point-cloud framework that relies on such rich 3D features.

The final loss function $\mathcal{L}_{total}$ is given by:
\begin{equation}
\small{
    \mathcal{L}_{total} =  \lambda_1\mathcal{L}_{CD} +   \lambda_2\mathcal{L}_{n} + \lambda_3\mathcal{L}_{l} + \lambda_4\mathcal{L}_{e} + \lambda_5\mathcal{L}_{c} +
     \lambda_6\mathcal{L}_{y}
     \label{eq:total_loss}}
\end{equation}

where $\lambda_1, \ldots, \lambda_6$ are training hyper-parameters. During inference, the encoder network takes as an input a single tongue image and predicts a 3D embedding $\mathbf{\tilde{y}}$, which is later transformed to the corresponding $\mathbf{p}_t$ parameters of the synthetic expression model, through the MLP between the two latent spaces. Finally, we apply the PCA model of the rigged head model on these $\mathbf{p}_t$ parameters to derive the final mesh of the head with the tongue expression. An overview of the methodology can be seen in Fig. \ref{fig:method}. 

\subsection{TongueGAN for ``in-the-wild'' reconstruction}
\label{sec:gan}
Although the pipeline presented in Section \ref{sec:method} provides a good estimation of the tongue pose in the test set of our collected data, it does not perform very well in ``in-the-wild'' images (Fig. \ref{fig:qual_itw_test}). This behavior is expected because our collected data were captured in controlled conditions and the training of the encoder was carried out only with rendered images which do not fully mimic ``in-the-wild'' conditions. To make our approach robust in ``in-the-wild'' images too, we would need to further train the pipeline by using also such data. However, for ``in-the-wild'' collected images from the web, we do not have their corresponding 3D tongue point-clouds. As a result, to use ``in-the-wild'' data in the pipeline, we would first need to have a method that can learn the distribution of our collected 3D tongue data and generalize well.

Finding a method to generate novel 3D tongues is tricky. This is because of several unique properties of the human tongue: a) it is a highly deformable object, so we cannot register our collected data in a reference template and apply relevant methods \cite{bouritsas2019neural,moschoglou20193dfacegan,potamias2020learning}, b) it is a non-watertight surface (i.e., it contains holes) so we cannot also use any implicit function approximations methods \cite{park2019deepsdf, saito2019pifu, michalkiewicz2019deep} or volumetric approaches \cite{jackson2017large,jackson20183d,zheng2019deephuman}. Therefore, having excluded the aforementioned categories, we decided to use GANs \cite{goodfellow2014generative} for the 3D tongue surface generations.

In order to generate accurate point-clouds that correspond to certain tongue images, our GAN, dubbed as TongueGAN, needs to be guided by meaningful labels which can capture all the desired 3D surface information. These labels are provided by the trained point-cloud AE as described in Section \ref{sec:tongue_recon}. Since the generation is driven by labels, TongueGAN is a conditional one \cite{mirza2014conditional}.

In particular, given a label denoted as $\mathbf{y}$ and a random Gaussian noise $\mathbf{z}\in\mathbb{R}^{128}$, the generator $G$ produces a novel point-cloud \textsl{point} $G\left(\mathbf{z}, \mathbf{y}\right)\in\mathbb{R}^{3}$, which we denote as $\mathbf{\tilde{x}}_t$, that belongs to the tongue surface represented by the label $\mathbf{y}$. On the other hand, the discriminator $D$ receives as inputs the label $\mathbf{y}$, a real point-cloud point $\mathbf{x}_t$ (which belongs to the tongue represented by the label $\mathbf{y}$) and the generator output $\mathbf{\tilde{x}}_t$ and tries to discriminate the fake (\ie, generated) from the real point. In the mathematical parlance, this can be described as:

\begin{align}
      \begin{array}{ll}
        \mathcal{L}_D = \mathbb{E}_{\mathbf{x}_t}\left[\log D\left(\mathbf{x}_t, \mathbf{y}\right)\right] - \mathbb{E}_{\mathbf{\tilde{x}}_t}\left[\log D\left(\mathbf{\tilde{x}}_t, \mathbf{y}\right)\right],\label{eq:adv_loss}\\
          \mathcal{L}_G = \mathbb{E}_{\mathbf{\tilde{x}}_t}\left[\log D\left(\mathbf{\tilde{x}}_t, \mathbf{y}\right)\right]
        \end{array}
\end{align}

where $D$ tries to maximize $\mathcal{L}_{D}$, whereas $G$ tries to minimize $\mathcal{L}_{G}$. The training process is considered complete when $D$ is no longer able to differentiate between the real and fake point-cloud points. 

Please note that instead of generating whole point-clouds for every provided pair $(\mathbf{z}, \mathbf{y})$ of noise and label, respectively, we merely generate a point corresponding to the surface which the label $\mathbf{y}$ represents. That confers several advantages in comparison to the rest of the methods in the literature, such as: a) we do not need to have in our training set point-clouds with the same number of points and as a result we can train our GAN without any data pre-processing on the raw point-clouds, which do not have a fixed number of points among them, b) when it comes to generating point-clouds corresponding to a particular label, we can generate on demand as many points as we want and, contrary to the rest of the literature, we are not constrained by any initially fixed resolution.

Since their inception, a plethora of GAN architectures and losses have been introduced in the literature \cite{arjovsky2017wasserstein, gulrajani2017improved, chen2016infogan, cheng2019meshgan, mao2017least}. For the TongueGAN loss we chose the Wasserstein loss with Gradient Penalty (WGAN with GP) \cite{gulrajani2017improved} due to its stability and good performance throughout the training process. As far as the architecture is concerned, we turned our attention to the recently proposed $\Pi$-Nets \cite{chrysos2020p, chrysos2020deep}, which are easy to implement and achieve state-of-the-art results in a large battery of tasks, including graph representation learning. In particular, we use our own, custom modification of $\Pi$-Nets to accommodate the needs of our task. A graphical presentation of the network structure is provided in Fig. \ref{fig:net_diagram}. 

\begin{figure}[t]
\begin{center}
\includegraphics[width=0.85\linewidth]{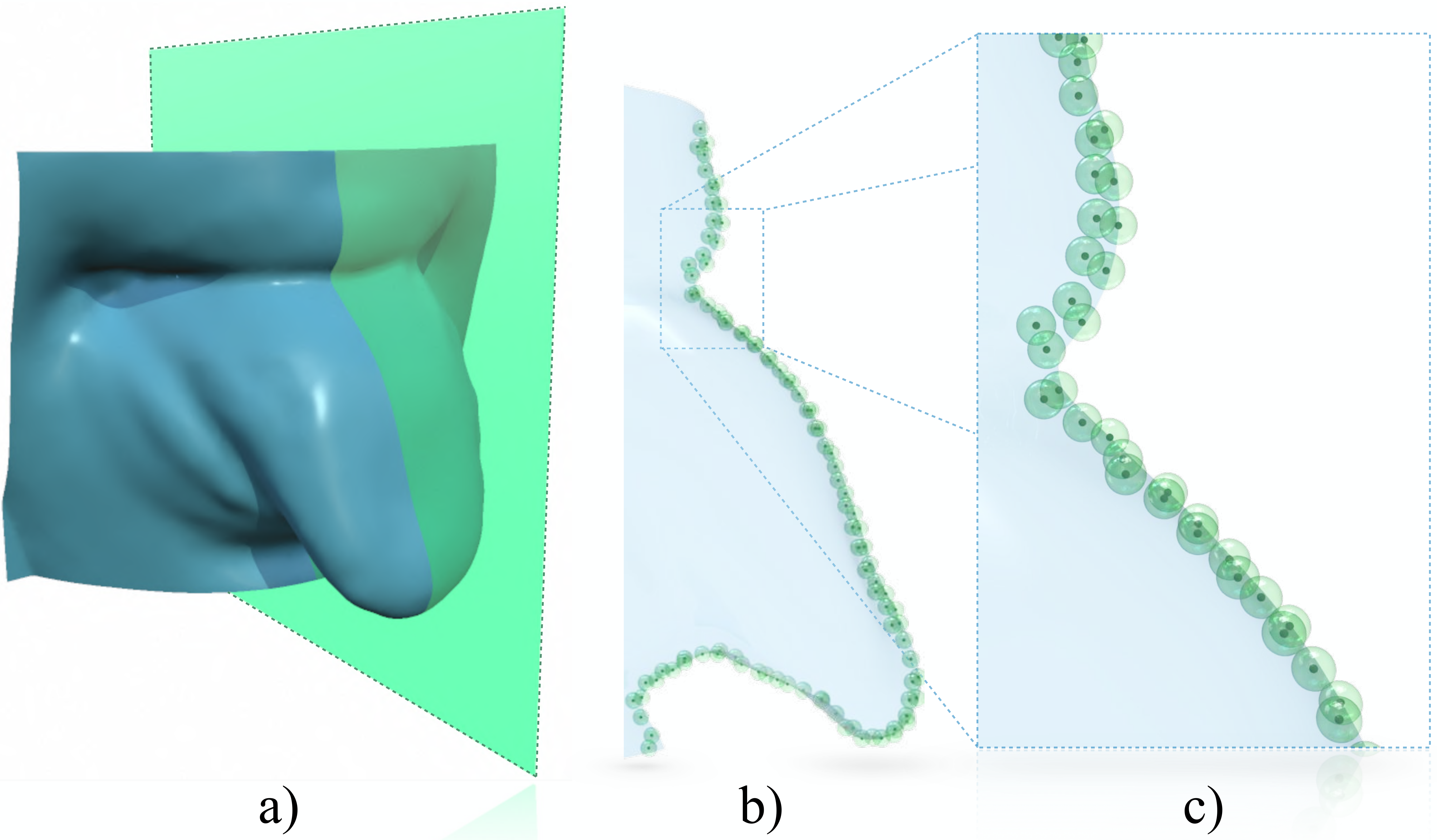}
\caption{A visual description of the mechanism explained in Section \ref{sec:novel_loss}. In a) we depict a raw tongue mesh with blue colour and with green colour we draw a section. In b) we show the 3D points (green color) that belong to the tongue surface and lie upon the section and with lighter green we depict the accepted areas from which points can be sampled and then fed to the discriminator as real data. These lighter green areas become gradually smaller as the training evolves till they collapse to the ground-truth 3D points. In c) we provide a zoomed-in area. }
\label{fig:margin_points}
\end{center}
\end{figure}

\subsubsection{GAN loss for accurate surface approximation}
\label{sec:novel_loss}

Even though, as can be clearly seen in the experiments, our custom $\Pi$-Nets modification along with the WGAN-GP loss significantly improve upon the vanilla GAN or point-cloud GAN \cite{li2018point}, there is still room for improvement as the point-cloud representations are not ideally reconstructed (see Table \ref{tab:errors}). 

We primarily attribute this to the strict behavior of the discriminator in GANs (\ie, deciding in our case whether a generated point matches exactly a point of the target point-cloud). This rigidity, especially in the early steps of the training process, is not very helpful, as the generator struggles to learn the real distribution of the point-clouds (\ie, all of the generated points are discarded as fake by the discriminator with high confidence). 

To remedy this, we slightly softened the discriminator, especially in the initial steps, by slightly modifying the real points fed to it. To achieve this, instead of directly feeding a real point $\mathbf{x}_t$ corresponding to a label $\mathbf{y}$ to the discriminator, we feed the following:

\begin{align}
\mathbf{x}_{t'|\mathbf{y}} \sim \mathcal{N}\left(\mathbf{x}_{t|\mathbf{y}}, \sigma_{e}\mathbf{I}
\right)\label{eq:sampling}
\end{align}

where $\mathcal{N}\left(\mathbf{x}_{t|\mathbf{y}}, \sigma_{e}\right)$ is a multi-variate normal distribution with mean $\mathbf{x}_t$ and (isotropic) variance $\sigma_e$. The variance $\sigma_e$ is not dependent on the label $\mathbf{y}$. It is only dependent on the epoch $e$. By employing \eqref{eq:sampling}, especially when the training process commences, the generator can better learn the actual distribution as it does not get severely punished by the discriminator when it slightly misses out the actual surface (see the accompanying Fig. \ref{fig:margin_points} for a visual understanding). As can be also seen in the experiments, this addition yields better results and stabilizes the training even further. We begin the training with a relatively small value for the variance and further linearly reduce it as we go along with the training process till it basically becomes zero towards the final epochs. This is further corroborated empirically in Section \ref{sec:experiments}.

\subsubsection{Re-training the pipeline}
\label{sec:retraining}
After the training is complete, we use the trained generator together with the trained encoder from Section \ref{sec:method}. In this way, we create ``in-the-wild'' pairs of 2D/3D data as follows: we feed the 2D ``in-the-wild'' image to the encoder and get the label $\tilde{{\mathbf{y}}}$. We then use this label $\tilde{{\mathbf{y}}}$
and the generator to produce a 3D point-cloud of the input image. As we can see in Fig. \ref{fig:itw_qual}, although TongueGAN is trained only on our collected data, it is able to generalize very well in ``in-the-wild'' images and as a result we can use it to create 2D/3D tongue pairs. We apply this process to a number of ``in-the-wild'' images to create multiple pairs. Finally, we re-train the pipeline we presented in Section \ref{sec:method}, using also the aforementioned pairs.

\section{Experiments}
\label{sec:experiments}
This Section is organized as follows. We begin by providing details regarding the training of the networks. Moreover, in Section \ref{sec:control}, we outline a series of quantitative as well as qualitative experiments under control conditions and finally, in Section \ref{sec:ITW} we describe our results under ``in-the-wild'' images.

The MLP utilized for the regression between the labels $\tilde{\mathbf{y}}$ and the PCA parameters $\mathbf{p}_t$ in Section \ref{sec:method} has a structure of (256, 128, 110) with a ReLU activation in the intermediate layers. The hyper parameters of $\eqref{eq:total_loss}$ which balance the losses are $\lambda_1=1.2$, $\lambda_2=1.6e-4$, $\lambda_3=0.4$, $\lambda_4=0.2$, $\lambda_5=0.8$ and $\lambda_6=1.5$. As described in Section \ref{sec:gan}, for TongueGAN we used a variant of WGAN with GP \cite{gulrajani2017improved}, which includes the injection mechanism \cite{chrysos2020p}, as well as the surface loss function presented in Section \ref{sec:novel_loss}. More specifically, we utilized a 9-layer Generator ($G$) and a 8-layer Discriminator ($D$) with a total number of parameters of about $8\times 10^6$ and $4\times10^6$, respectively. We trained TongueGAN using the Adam optimizer \cite{kingma2014adam} with $\left(\beta_1=0,\beta_2=0.9\right)$. We also trained with a batch size of $2048$ for a total of $10^6$ iterations. Following the idea introduced in \cite{heusel2017gans}, we use individual learning rates for D and G with values of $1e-4$ and $1e-5$, respectively. Finally, we start training with the variance $\sigma_e$ in $\eqref{eq:sampling}$ being $5e-3$ and we linearly decrease it by $10\%$ every $50\times10^3$ steps. The exact network structures are deferred to the supplementary material with more details. 

\subsection{3D tongue reconstruction in control conditions}
\label{sec:control}
In this set of experiments, we used $90\%$ of TongueDB for training and the rest for testing. Due to the intricacy of the tongue as a surface (as we explained in detail in Section \ref{sec:gan}), we decided to use a GAN for the tongue surface generation part. Moreover, an extra reason to utilize a GAN for the training is the fact that it is able to generalize very well in unseen labels during testing. To the best of our knowledge, the only method which has been introduced in the literature and is able to carry out point-cloud generations based on unseen labels is PointCloud GAN (PC-GAN) \cite{li2018point}. Consequently, in what follows we draw comparisons against PC-GAN \cite{li2018point} and another two variants of TongueGAN, namely: a) TongueGAN\_v1, which is the same as TongueGAN with the only difference being that the novel loss function (Section \ref{sec:novel_loss}) is not available in this version, and b) TongueGAN\_v2, which is the typical GAN structure where, instead of the injections we have simple concatenations along the layers. Finally, we also report the results for the regressed tongue expression (referred to as Tongue-Reg). For this, we only take into account a small patch around the oral cavity defined by the ground truth point cloud, in order to deduce a reasonable error. 

\begin{figure}[t]
\begin{center}
\includegraphics[width=0.95\linewidth]{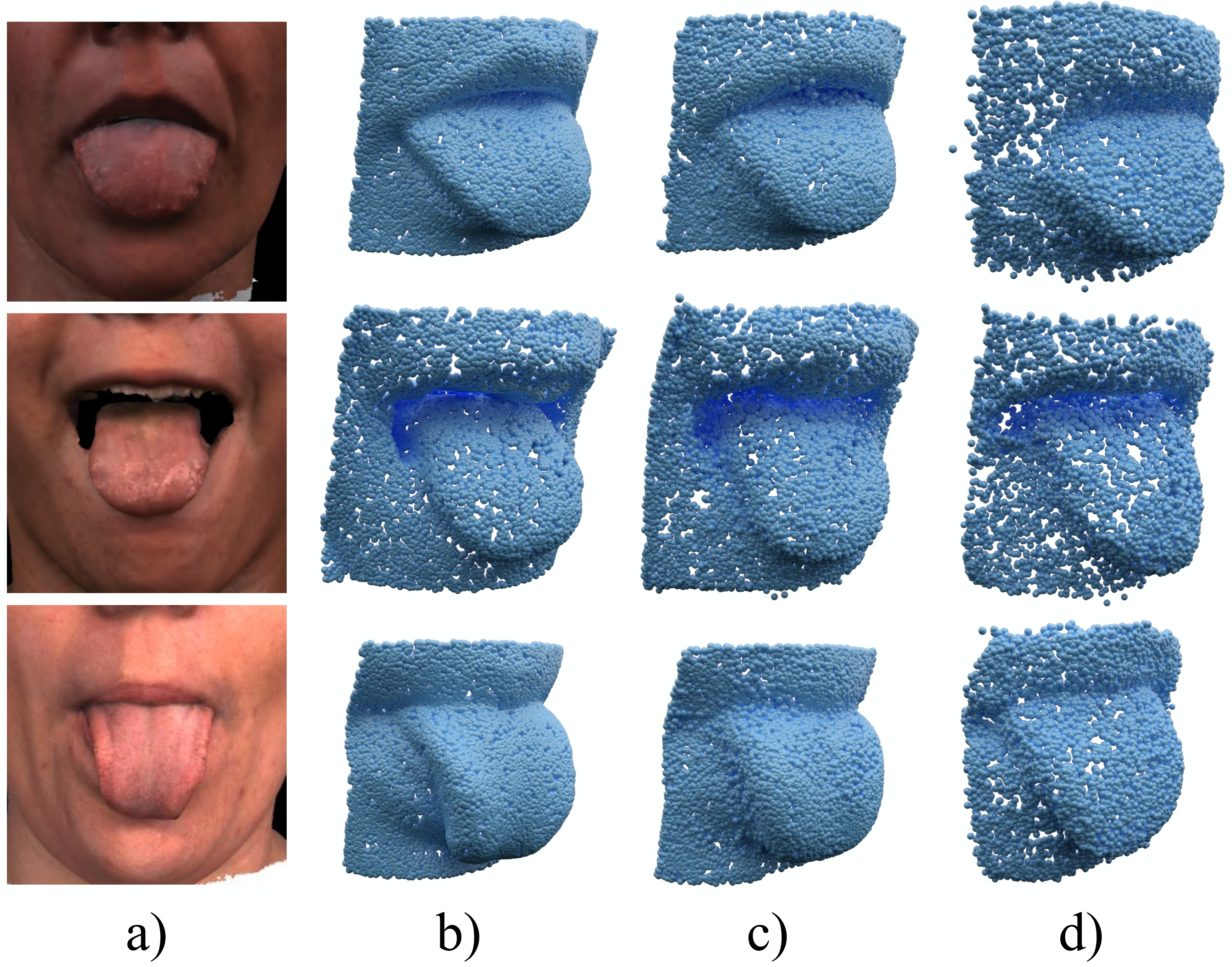}
\caption{Qualitative comparisons on various pointclouds from the test set of TongueDB. a) input image, b) ground-truth point-cloud, c) the point-cloud generated by TongueGAN and finally d) the point-cloud generated by point-cloud GAN \cite{li2018point}.}
\label{fig:point_cloud_comp}
\end{center}
\end{figure}


Quantitative results are provided in Table \ref{tab:errors} and qualitative results are presented in Fig.\ref{fig:point_cloud_comp}. For the quantitative results, we utilize the test set of our TongueDB and we measure the error based on the two commonly used type of distances when it comes to unordered 3D data, namely Chamfer Distance (CD) and Earth Mover's Distance (EMD) \cite{fan2017point}. As can be clearly seen in all of the comparisons, TongueGAN outperforms the compared methods by a large margin whereas the regressed tongue expression outperforms the rest of the methods. 

\begin{figure*}[t]
\begin{center}
\includegraphics[width=1\linewidth]{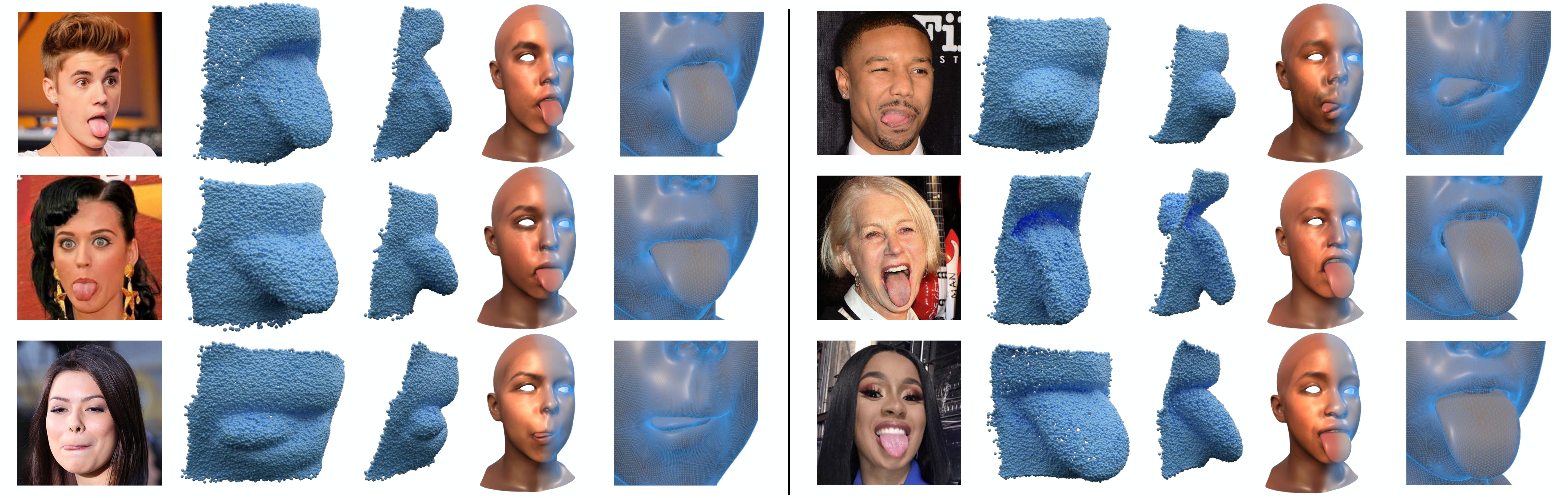}
\caption{3D head reconstructions with tongue animations from ``in-the-wild'' images. From left to right, we depict the ``in-the-wild'' image, then the point-cloud generations from two viewpoints and finally the 3D head reconstruction with a zoomed-in area around the oral cavity.}
\label{fig:itw_qual}
\end{center}
\end{figure*}

\begin{table}[t]
\caption{Quantitative comparisons among the compared methods using CD and EMD as metrics. Lower values indicate better performance. TongueGAN achieves the best results in all settings.}
\label{tab:errors}
\vspace{0.3cm}
\centering
\begin{tabular}{|l|cc|}
\hline
\emph{Method} & \emph{EMD} & \emph{CD} \\
\hline\hline
\textbf{TongueGAN} & \textbf{1.62e-2} & \textbf{5.25e-5} \\
Tongue-Reg & 1.79e-2 & 1.10e-4\\
PC-GAN & 1.82e-2 & 1.13e-4 \\
TongueGAN\_v1  & 1.97e-2 & 1.67e-4 \\ 
TongueGAN\_v2 & 2.24e-2 & 2.09e-4 \\
\hline
\end{tabular}
\end{table}


\subsection{3D tongue reconstruction ``in-the-wild''}
\label{sec:ITW}
In this Section, we attempt to reconstruct the 3D surface of the tongue together with the entire head structure from ``in-the-wild'' images. In this set of experiments, we used all of TongueDB for training. We also added to our training data another $5K$ ``in-the-wild'' tongue images and created their 3D point-clouds using TongueGAN. Using all these data, we re-trained the pipeline according to Section \ref{sec:retraining}. The results are only visual as we do not have ground-truth point-cloud data to report quantitative comparisons. 
Regarding the comparisons, we should note that PointCloud GAN \cite{li2018point} cannot be used in these experiments, as in order to work in the conditional setting, it needs as input the actual ground-truth point-cloud it attempts to reconstruct, something which we do not have at our disposal. Given that the TongueGAN variations (\ie, TongueGAN\_v1 and TongueGAN\_v2) perform worse than PointCloud GAN \cite{li2018point}, we only present results in this Section regarding TongueGAN. 

Since our tongue regression method is based on the mean template mesh of UHM we can easily utilize the pipeline presented in \cite{ploumpis2020towards} in order to extent our approach to a particular facial identity. We begin by fitting a facial mesh to the image domain in order to get the 2D/3D landmarks and the identity of the subject and then we regress to the full head topology based on the UHM model. After reconstructing the head shape we crop the image around the projected 2D mouth landmarks. We then feed this cropped image to the re-trained pipeline and get the mean head shape with the tongue expression as mentioned in Section \ref{sec:tongue_recon}. Finally we merge the predicted tongue shape with the associated identity by treating the predicted tongue expression as a separate blend-shape.

\begin{figure}[t]
\begin{center}
\includegraphics[width=0.95\linewidth]{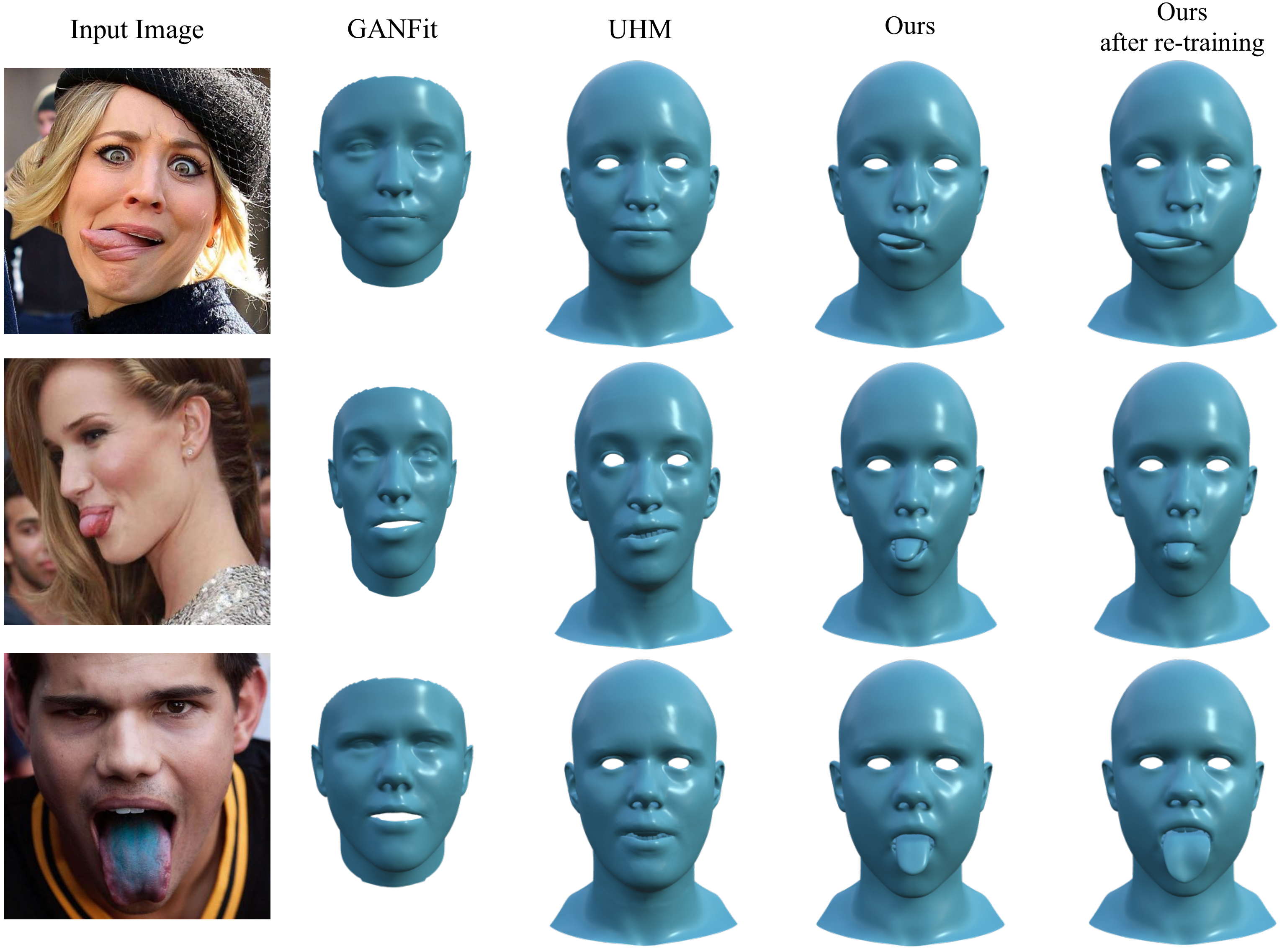}
\caption{Qualitative shape evaluation between our approach and the state-of-the-art methods of facial \cite{gecer2019ganfit} (GANFit) and head \cite{ploumpis2020towards} (UHM) reconstructions. We can easily deduce that the re-training framework plays an important role in the final tongue reconstruction from ``in-the-wild'' images.}
\label{fig:qual_itw_test}
\end{center}
\end{figure}

Some 3D reconstructions can be seen in Fig. \ref{fig:itw_qual}. As evidenced, our pipeline is able to accurately reconstruct the 3D tongue details even in ``in-the-wild'' conditions. Additional tongue reconstructions of our method before and after the re-training framework against state-of-the-art methods can be seen in Fig. \ref{fig:qual_itw_test}. To further empirically validate that TongueGAN is able to capture the 3D structures of random tongues that are not included in the training set we provide linear interpolations between unseen latent features in the supplementary material.


\section{Conclusion}
\label{sec:conclusions}
In this work, we presented the first pipeline which is able to perform 3D head and tongue reconstruction from a single image. To achieve this, we collected the first diverse tongue dataset with various tongue shapes and positions which we make publicly available to the research community. To also make this pipeline robust in ``in-the-wild'' images and to mitigate the absence of their corresponding ground-truth 3D tongue data, we introduced the first GAN method that is tailored for accurately reconstructing the 3D surface of a tongue from 2D images. As we show in a series of experiments, we are now able to accurately carry out 3D head reconstruction together with the tongue from a single image and thus create more realistic 3D avatars. 

{\small
\bibliographystyle{ieee_fullname}
\bibliography{egbib}
}

\end{document}